# Large Language Models for Conducting Advanced Text Analytics Information Systems Research

Benjamin M. Ampel

Georgia State University, bampel@gsu.edu

Chi-Heng Yang

University of Arizona, chihengyang@arizona.edu

James Hu

University of Arizona, jameshu@arizona.edu

Hsinchun Chen

University of Arizona, hsinchun@arizona.edu

The exponential growth of digital content has generated massive textual datasets, necessitating the use of advanced analytical approaches. Large Language Models (LLMs) have emerged as tools that are capable of processing and extracting insights from massive unstructured textual datasets. However, how to leverage LLMs for text analytics Information Systems (IS) research is currently unclear. To assist the IS community in understanding how to operationalize LLMs, we propose a Text Analytics for Information Systems Research (TAISR) framework. Our proposed framework provides detailed recommendations grounded in IS and LLM literature on how to conduct meaningful text analytics IS research for design science, behavioral, and econometric streams. We conducted three business intelligence case studies using our TAISR framework to demonstrate its application in several IS research contexts. We also outline the potential challenges and limitations of adopting LLMs for IS. By offering a systematic approach and evidence of its utility, our TAISR framework contributes to future IS research streams looking to incorporate powerful LLMs for text analytics.

## 1 INTRODUCTION

The proliferation of digital content (social media, online reviews, blogs, etc.) has provided massive quantities of unstructured textual data for organizations and researchers to derive new and interesting insights [12]. Organizations and researchers are increasingly relying on artificial intelligence (AI)-enabled methodologies to automatically analyze large textual datasets [13]. Deep learning approaches are particularly well suited for large-scale text analytics tasks as they automatically discover and engineer latent features within textual datasets [77]. Information systems (IS) researchers have demonstrated the value of deep learning frameworks for a wide range of advanced text analytics applications, including healthcare [86], cybersecurity [3], and FinTech [21]. These deep learning frameworks have enabled fine-grained insights that were previously non-trivial to obtain.

Although these deep learning frameworks provide tremendous value, they have several limitations. Prior IS literature initially focused on long short-term memory (LSTM) models to conduct large-scale text analytics [3, 86]. LSTMs are designed for sequential processing (i.e., processing one token at a time). Sequential processing introduces substantial computational overhead, struggles to efficiently capture long-term dependencies, and prevents parallelization. These limitations make LSTMs impractical for large-scale text analytics. Large Language Models (LLMs) have emerged as a potential solution to the limitations of sequential models in text analytics IS research. LLMs are advanced deep learning models that are pre-trained on massive amounts of textual data and then fine-tuned for downstream tasks [76]. More recently, IS literature has adapted LLMs for social bot detection [7], review classification [18], cybersecurity linking [4], and resilience detection [59]. These studies demonstrated the significant promise of LLMs for text analytics IS research. However, these studies implemented LLMs for classification, which is not representative of the various tasks that LLMs can perform in IS research (e.g., summarization and generation tasks). LLMs are particularly well suited for summarization and generation tasks because of their significant pre-training [79, 90].

To assist IS researchers in operationalizing LLMs, we propose a Text Analytics Information Systems Research (TAISR) framework. Such frameworks are useful to the IS community when commonly accepted guidelines do not exist [12]. The proposed TAISR framework builds upon existing recommendations for conducting AI-enabled IS research by providing detailed information about LLMs from technical and implementation perspectives [63, 70]. Our framework also differentiates itself from extant surveys on LLMs by being application-driven and targeted for the field of IS [33, 51, 57]. While this study is not comprehensive of LLMs, it provides carefully synthesized recommendations and contributions that IS researchers can build upon. This study makes three contributions to the IS knowledge base [29]:

1. An overview of LLMs as a guide to IS researchers unfamiliar with the low-level technical details of these models. This overview covers several technical concepts to facilitate the understanding and operationalization of LLMs for text analytics IS research.
2. The TAISR framework, which builds upon extant IS research frameworks and computer science literature to provide a clear set of guidelines for defining the research objective, choosing an appropriate text analytics task, collecting relevant data, implementation, and evaluating the LLM. Although many TAISR concepts have been drawn from IS research frameworks [70], TAISR provides recommendations that are applicable only to LLMs.
3. Three targeted case studies, which demonstrated the proof-of-value and proof-of-concept of the TAISR model [60]. These case studies are potentially generalizable to several domains and can be built upon by future IS researchers. As a result, these case studies can rapidly advance IS text analytics research.

The remainder of this paper is organized as follows. First, we provide an overview of LLMs. Second, we summarize the current seminal LLMs by closed-source and open-source. Third, we introduce our LLM for TAISR framework. Fourth, we provide three case studies that show the proof-of-value of our proposed TAISR framework. Fourth, we discuss the current challenges and limitations of LLM implementation. Finally, we conclude this paper.

## 2 AN OVERVIEW OF LARGE LANGUAGE MODELS

The first LLMs adapted the bidirectional LSTM architecture and pre-trained on a corpus of 30 million sentences [65]. These recurrent-based LLMs improved over traditional embedding approaches (e.g., Word2Vec and GloVe) by learning word contexts based on the sentence around each word. Word contexts were learned through language modeling, which is a self-supervised approach for predicting the next word in a sentence based on the prior and future contexts of the word. However, researchers have noted that recurrent architectures are not scalable due to vanishing gradients (gradients diminish exponentially during backpropagation) and sequential processing [57].

To alleviate vanishing gradients and sequential processing issues, LLM researchers maintained the same language modeling training procedure (self-supervised next-word prediction) but replaced the underlying recurrent-based architecture with the newer transformer architecture [80]. The transformer operates via a self-attention mechanism that concurrently attends to each token in the input sequence. Concurrent processing enables the model to process sequences in parallel, rather than sequentially. Therefore, the self-attention mechanism alleviates both vanishing gradient and scalability concerns while capturing contextual dependencies between tokens.

The original transformer architecture is a sequence-to-sequence model comprising an encoder and a decoder [80]. The encoder and decoder function as distinct deep learning models designed to learn latent representations from input sequences. The encoder consists of a multi-head attention block (several layered self-attention mechanisms) and a feed-forward layer. The decoder employs a masked multi-head attention block (which limits the decoder from seeing future context) to process inputs. This process ensures that the decoder effectively learns the latent representation mapping input (from the encoder) to the output (from the decoder) sequences.

Similar to recurrent-based LLMs, transformer-based LLMs often use self-supervised pre-training (i.e., language modeling) using massive textual datasets comprising web text (e.g., Wikipedia) and books [67]. To perform pre-training, the text was parsed into individual sentences. Tokens within the text are masked (which can be random or at the end of the sentence based on LLM architecture) [20]. The pre-training objective aims to maximize the conditional probability of predicting masked tokens given unmasked tokens. A pre-trained LLM can then be used (i.e., zero-shot) or adapted for specific downstream tasks. Through the careful combination of the transformer architecture and pre-training, LLMs have attained state-of-the-art performance in nearly every natural language processing (NLP) task [79].

The most common adaptation method for LLMs is fine-tuning. During fine-tuning, the weights of the LLM are adjusted based on a task-specific dataset [57]. For example, fine-tuning an LLM for review sentiment classification could include updating the model parameters using a dataset of labeled review-sentiment pairs. Fine-tuning strategies include full and partial fine-tuning. Full fine-tuning updates all layers in an LLM and should be implemented when the target task is significantly different from the LLM's pre-training objective [57]. In partial fine-tuning, only specific layers are modified (e.g., the final linear layer for a classification task) [72]. Partial fine-tuning is preferable when the pre-trained model has already captured relevant features for the target task in its lower layers and modifications to higher layers are sufficient for optimal task performance. The current salient methods of partial fine-tuning are low-rank adaptation (only a subset of important parameters is adjusted) and adapter layers (small trained layers that specialize in a chosen task) [27].

Reinforcement learning is an increasingly popular form of fine-tuning [62]. Reinforcement learning is a class of methods that aims to align LLM outputs towards a specific goal using policy updates. For example, research has used Proximal Policy Optimization (PPO) to guide LLM-generated text towards specific sentiments [62]. For PPO-based reinforcement learning, the LLM generates words or sequences based on its current state and policy, receiving rewards or scores based on how well its actions align (e.g., generating the correct sentiment) with the desired outcomes. Another type of reinforcement learning, known as reinforcement learning from human feedback (RLHF), aims to align LLM outputs with human preferences. RLHF, which is implemented in ChatGPT, uses manually annotated rankings of model outputs to create a reward model that guides future model generation [61, 62]. The nascent area of reinforcement learning for LLM fine-tuning should be carefully monitored by IS researchers for future updates and potential novel contributions.

Transformer-based LLMs can be grouped into three categories: (1) encoder, (2) encoder-decoder, and (3) decoder. We visualize each of the three categories of LLMs in Figure 1 and then detail each in the following sections.

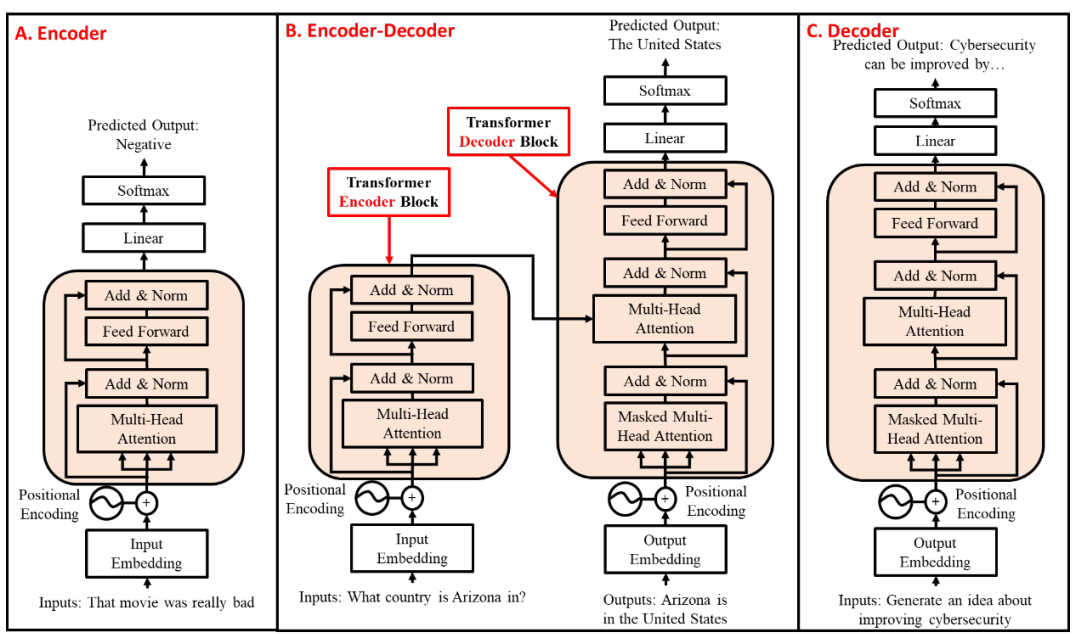

Figure 1: The Three Types of LLM Architectures: (A) Encoder, (B) Encoder-Decoder, and (C) Decoder

## 2.1 Encoder LLMs

Encoder LLMs (e.g., BERT) (Figure 1A) are most often trained self-supervised using masked language modeling [57]. In masked language modeling, 15% of the tokens in an input sentence are randomly masked (i.e., hidden) from the encoder LLM. A softmax layer is placed at the end of the encoder LLM to predict the masked token(s). The goal of an encoder LLM is to maximize the log-likelihood function:

$$L(X) = \sum_{i-1} log P([Mask_i] = y_i | \tilde{X}; \Theta) \qquad (1)$$

where $X$ is the corpus of tokens, $Mask_i$ is a token that we want to predict ($y_i$) given the context tokens of $\tilde{X}$, and $\Theta$ are the parameters learned by the LLM. Some encoder architectures also include a simultaneous next-sentence prediction task [20]. In the next-sentence prediction, the training dataset is randomly split into sentence pairs that follow each other in the corpus and pairs that do not. The training task is a simple binary classification (does the second sentence follow the first sentence? Yes/No). This process is illustrated in Figure 2.

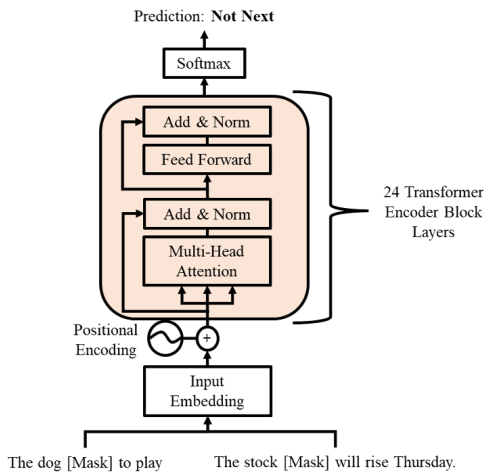

Figure 2: Encoder Next Sentence Prediction (Adapted from [20])

Although encoder LLMs can be used for language generation tasks (e.g., summarization), they are better suited for classification owing to their masked language modeling training procedure [57]. Fine-tuning an encoder LLM for classification tasks is the same as training any supervised model on a task-specific dataset (i.e., the LLM is provided a set of inputs and class labels). Therefore, the metrics used to evaluate the outputs of encoder LLMs are the same as standard classification tasks: accuracy, precision, recall, F1-score, or area under the curve (AUC) [42].

## 2.2 Encoder-Decoder LLMs

Encoder-decoder LLMs (e.g., BART) (Figure 1B) employ both the encoder and decoder of the standard transformer. Encoder-decoder LLMs are often trained using denoising autoencoding [46]. In denoising autoencoding, text is corrupted through token masking and token shuffling to train the model on both masked and permutation language modeling objectives. The model aims to reconstruct the original sequence from the corrupted version. Encoder-decoder LLMs often use the following loss function:

$$\mathcal{L}_{DAE}(X, X') = -\log P(X|X') \qquad (2)$$

where $\mathcal{L}_{DAE}$ is the denoising autoencoding loss function and $P(X|X')$ is the conditional probability of generating the original sequence $X$ given a corrupted sequence $X'$ [46]. This training approach enables the encoder-decoder LLM to capture contextually rich textual representations while excelling in tasks such as text summarization and translation [89].

An important facet of encoder-decoder (and decoder) LLMs is the choice of decoding method. A decoding method is a post hoc process of converting the output of an LLM's internal representation into a readable sequence of text [31]. There are three types of decoding: (1) deterministic, (2) stochastic, and (3) contrastive. First, deterministic methods (e.g., greedy and beam search) select the next token based on maximizing the probability distribution of the model [26, 45]. Greedy search chooses the most probable token provided by the LLM [45]. Beam search calculates the sequence log probabilities to select the most probable sequence [26]. Deterministic methods often lead to high coherence but low diversity due to repeating high probability $n$-grams (known as neural degeneration, shown in the top left of Figure 3). Deterministic methods should be used when users want a text summary to contain high-probability tokens from a document. Second, stochastic methods (e.g., top-$k$ sampling and nucleus sampling) aim to address neural degeneration by truncating the probability distribution tail and sampling the remaining tokens (shown on the right side of Figure 3) [31]. Top-$k$ sampling samples $k$ (set by user) probable tokens and chooses one [84]. Top-$k$ samples the same $k$ for each token in a sequence, which is undesirable when the probability distribution varies significantly from token to token. Nucleus sampling samples from a set of tokens that add up to a cumulative probability (set by the user) [31]. Stochastic methods often lead to diverse sentences but low semantic consistency. Stochastic methods should be used when a user wants more variability (e.g., idea generation). The token selection process for greedy, beam, top-$k$, and nucleus sampling is illustrated in Figure 3.

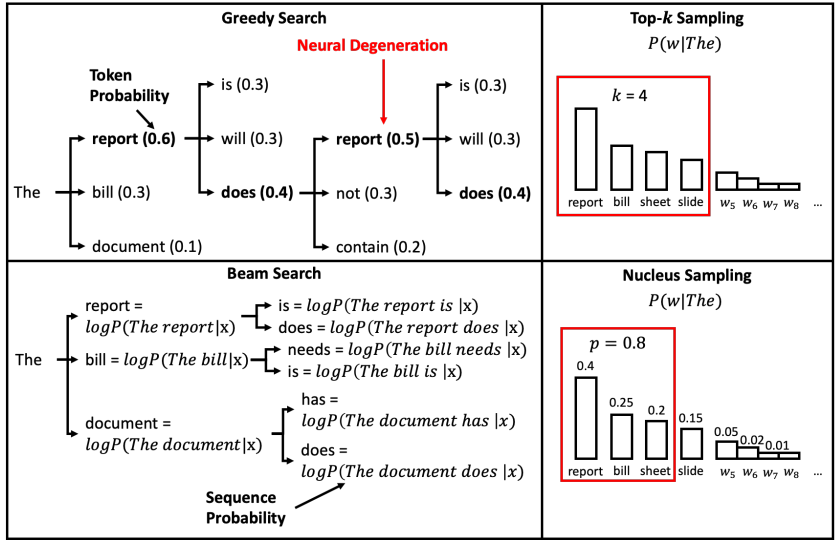

Figure 3: Token Selection Process for the Four Most Popular Decoding Methods

Finally, contrastive search is a new form of decoding that can be used to achieve a balance between coherence and diversity [75]. Contrastive search uses the contrastive objective function:

$$x_t = \underset{v \in V^k}{arg\,max} \left\{ (1 - \alpha) \times p_\theta(v|x_{<t}) - \alpha \times \left( max \left\{ s\left(h_v, h_{x_j}\right) : 1 \leq j \leq t - 1 \right\} \right) \right\}, \qquad (3)$$

where $p_\theta(v|x_{<t})$ is the LLM's probability (i.e., confidence) that candidate token $v$ follows given the prior tokens $x_{<t}$ at time $t$ and $max \left\{ s\left(h_v, h_{x_j}\right) : 1 \leq j \leq t - 1 \right\}$ is a degeneration penalty that measures the cosine similarity between $h_v$ and $h_{x_j}$.

Regardless of the decoding strategy, the evaluation of encoder-decoder LLMs used for text generation or summarization can be automated or human-annotated based on the goal of the user [89]. Automated metrics can be reference-free (does not require ground-truth text) or reference (requires ground-truth text). We provide seminal metrics (including their pros and cons) to evaluate the natural language outputs of an LLM in Table 1. The choice of evaluation metric(s) should be based on the task and access to ground truth data or human annotators.

Table 1: A Summary of Evaluation Metrics for Generative LLMs

| Evaluation | Type | Metric | Explanation | Pro(s) | Con(s) | Reference |
|---|---|---|---|---|---|---|
| Automated | Reference-free | Coherence | Measures the coherence between generated and prefix text using an LLM | Correlates well with human evaluations | Chosen LLM may suffer from inductive bias | [75] |
| | | Diversity | Measures token repetitiveness | Easy to calculate | Not informative of text quality | [84] |
| | | MAUVE | Measures the KL divergence between neural and human text with an LLM | Great for relative comparisons | Chosen LLM may suffer from inductive bias | [66] |
| | | Perplexity | Likelihood the model generated the input text sequence | Can determine if the LLM learned the text distribution | Not comparable between models or datasets | [84] |
| | Reference | BERTScore | Evaluates the similarity of two texts based on similarity of their BERT embeddings | Prevailing machine translation metric | Computationally expensive to calculate | [91] |
| | | BLEU | Evaluates the similarity of texts based on n-grams | Good for translation tasks | Rarely correlates with human evaluations | [64] |
| | | ROUGE | Evaluates the similarity of texts based on n-grams | Good for summarization tasks | Rarely correlates with human evaluations | [50] |
| | Hybrid | BARTScore | Likelihood that BART would generate a text based on references | Prevailing text summarization metric | Assumes BART is a good indicator of text quality | [89] |
| Human | N/A | Human likeness | Likelihood the given passage is AI-generated | Human evaluations are useful for automated comparisons | Humans are poor at detecting LLM text | [15] |
| | | Fluency | Measures if the text is easy to understand | | Subjective, costly to obtain | [75] |
| | | Informativeness | Measures if the text is interesting | | | |

Note: BLEU = Bilingual Evaluation Understudy, MAUVE = Mean Area Under the Divergence Curve Evaluation, ROUGE = Recall-Oriented Understudy for Gisting Evaluation.

Automated evaluation generally evaluates the textual outputs of LLMs based on statistical properties (e.g., $n$-grams), closeness with a ground-truth dataset (e.g., BLEU, ROUGE), or LLM evaluation (e.g., BARTScore). Reference metrics (BERTScore, BLEU, and ROUGE) can only be used when a ground-truth dataset exists. However, reference-free metrics can be used to calculate the statistical properties of text without a ground-truth dataset. Human evaluation metrics (Human Likeness, Fluency, and Informativeness) capture subjective qualities such as the text's human-like nature, ease of understanding, and intriguing content. Typically, these metrics are derived from a 5-point Likert scale [75]. These metrics are costly to obtain but can provide tremendous insight into the quality of LLM-generated text [75].

### 2.3 Decoder LLMs

Decoder LLMs (e.g., GPT-2 and ChatGPT) (Figure 1C) are primarily trained through autoregressive training. Autoregressive training is a self-supervised approach in which the model predicts the likelihood of the next token given the preceding context [68]. The decoder LLM processes each token in a sequence and generates subsequent token probabilities based on the preceding tokens using an autoregressive mechanism. The goal of autoregressive training is to maximize the log-likelihood of the entire sequence:

$$L_1(\mathcal{U}) = \sum_i log P(u_i | u_{i-k}, \dots, u_{i-1}; \Theta), \qquad (4)$$

where $\mathcal{U}$ is the textual corpus, $u_i$ is a token that we want to predict given the prior context tokens of $u_{i-k}$ and $u_{i-1}$, and $\Theta$ are the parameters learned by the decoder LLM.

Decoder LLMs typically do not use a standard row of data for input (like many encoder and encoder-decoder LLMs). Instead, seminal decoder LLMs often use a prompt as input [62]. Prompts can be basic or specific (e.g., provided contexts or constraints) and will result in vastly different outputs from the LLM. Prompt engineering allows users of an LLM to further control text generation [51]. Users provide a prefix to their prompt specifying the task (e.g., summarize the following passage) before providing an input (e.g., a passage). For example, chain-of-thought prompting (a type of prompt engineering) provides a series of reasoning examples to an LLM to achieve a target task [82]. Chain-of-thought prompting has been shown to improve model performance in multi-step tasks (e.g., complex math problems) and output interpretability (i.e., LLMs provide step-by-step logic to the user). As previously stated, these prompts can also be learned through a training strategy [51].

Like encoder-decoder LLMs, decoder LLMs rely on a decoding strategy (deterministic, stochastic, contrastive) to transform model probability distributions into natural language. The same recommendations outlined in the encoder-decoder subsection for choice of decoding strategy apply. Further, since decoder LLMs generate natural language text, the evaluation metrics and recommendations outlined in Table 1 are also applicable.

### 3 MAJOR PLAYERS IN LLMS

To help understand how LLMs are currently used in organizational contexts, we reviewed the organizational landscape by model access, company, flagship model(s), number of parameters, and language support in Table 2. We then further detail closed- and open-source models in subsequent sections.

Table 2: Current Seminal Closed- and Open-source Companies

| Access | Company | Flagship Model(s) | Parameters | Language Support |
|---|---|---|---|---|
| Closed-Source | OpenAI | GPT-4o | 1 trillion | 50+ natural languages |
| | Microsoft | Bing, GitHub CoPilot | 1 trillion | 11 programming languages |
| | Alphabet (Google) | Bard, Gemini, LaMDA | 137 billion | 50+ natural languages |
| | Anthropic | Claude 3 | 52 billion | English |
| Open-Source | Huawei | PanGu-$\alpha$, PanGu-$\Sigma$ | 1.1 trillion | ChatBot |
| | Meta | OPT, Llama3 | 175 billion | English |
| | Alphabet (Google) | Gemma, CodeGemma | 7 billion | English |

| Stanford | Alpaca | 7 billion | English |
|---|---|---|---|
| **Baidu** | Ernie 3.0 Titan | 260 billion | English |
| **EleutherAI** | GPT-Neo X | 20 billion | English |
| **Tsinghua University** | GLM | 10 billion | English, Chinese |
| **Technology Innovation Institute** | Falcon | 40 billion | English, Arabic |
| **BigScience** | BLOOM | 176 billion | 46 natural languages |

Closed-source LLMs primarily come from major technology companies (e.g., OpenAI, Microsoft, and Alphabet). These LLMs have rapidly increased their parameters to over one trillion and support many different natural or coding languages. Open-source LLMs are often produced by research universities (e.g., Stanford and Tsinghua) and have fewer parameters than flagship closed-source LLMs. Open-source LLMs often only support one or two languages. Open-source models often perform worse than open-source models in NLP tasks but provide significantly more access to the LLM architecture for careful control over model outputs [79].

### 3.1 Closed-source LLMs

Closed-source LLMs are LLMs developed by companies that choose to retain exclusive control over LLM design, training methodologies, and data. Users typically interact with these models through APIs or interfaces, leveraging their functionalities in various NLP tasks. We summarize the current closed-source LLMs by company, LLM purpose, API functions, inputs, outputs, and whether they can be fine-tuned in Table 3.

Table 3: A Summary of Current Seminal Closed-Source LLMs

| Company | Model | Purpose | API Functions | API Input | API Output | Fine-tuning? |
|---|---|---|---|---|---|---|
| **OpenAI** | **GPT-4o** | NLP, image, and voice tasks | Fine-tuning | Training data | Task completion | Yes |
| **GitHub** | **CoPilot** | Code completion | Coding tasks | Text/Code | Code block | Yes |
| **Alphabet (Google)** | **Gemini** | Chatbot with search capability | Chat, search | Search | Response, search results | No |
| | **PaLM** | NLP tasks | Fine-tuning | Training data | Task completion | Yes |
| | **Codey** | Code completion | Coding tasks | Text/Code | Code block | No |
| **Anthropic** | **Claude 3** | Ethics driven chatbot | Chat | Prompt text | Response | No |
| **Huawei** | **PanGu-α** | Chinese NLP tasks | Fine-tuning | Training data | Task completion | Yes |
| | **PanGu-Σ** | Chinese NLP tasks | Fine-tuning | Training data | Task completion | Yes |

Closed-source models offer a wide spectrum of applications, ranging from general-purpose chatbots (ChatGPT, LaMDA) to specific tasks, such as code completion (CoPilot, Codey), smart search engines (Bing, Bard), and ethics-driven chatbots (Claude). Inputs are often prompt text. The outputs are responses to these prompts, including completed code blocks and textual task completions. However, the closed-source nature of these LLMs introduces challenges related to transparency (unclear what data was used to train LLM), interpretability (difficult to ascertain why certain outputs were achieved), and accessibility (cannot choose decoding method) compared to their open-source LLM counterparts. LLMs supporting fine-tuning provide users with the flexibility to adapt and optimize performance for specific tasks or datasets. However, these LLMs may still have significant security concerns (e.g., logging proprietary input data). This underscores the trade-off between closed-source LLM functionalities and their potential limitations in academic research.

### 3.2 Open-source LLMs

Open-source LLMs are publicly available models that allow users to access their architecture, codebases, and training data. Open-source LLMs can be downloaded and adapted to perform a task using fine-tuning, prompt learning, or another adaptive strategy. Hugging Face is a salient Python library for training and operationalizing open-source LLMs [85]. Hugging Face provides access to over 163,000 LLMs, 26,000 datasets, numerous fine-tuning, prompting, and decoding strategies, tutorials, and other tools to quickly develop cutting-edge text analytics projects. Hugging Face's usefulness is primarily in training from scratch or implementing an existing open-source LLM to provide an output given an input.

Hugging Face alone does not provide sufficient tools to create powerful applications. For example, LLMs are not proper knowledge bases (often providing incorrect information to users) [37]. To incorporate knowledge bases, data, and other information into a fine-tuned LLM, additional tools are required. Several tools (e.g., LangChain) allow researchers to create multi-step applications powered by LLMs [78]. LangChain provides tools for building chatbot UIs, sequential processes (e.g., making a Google API request and prompting an LLM based on the results of the request), evaluations (e.g., determining whether a response is a good answer to a question using an ensemble of LLMs), and custom agents (e.g., an LLM can take actions based on observations).

LLM architectures are often built on a transformer encoder, encoder-decoder, or decoder [57]. We provide a list of seminal LLMs for each architecture with the estimated disk space and computational power to train each in Table 4.

Table 4: A Summary of Current Seminal Open-Source LLMs

| Architecture | Prevailing LLMs | Aggregate Disk Space | Estimated GPU |
|---|---|---|---|
| Encoder | BERT | 2,048 MB | 1.18 GB |
| | ALBERT | 1,248 MB | 0.19 GB |
| | XLM | 5,200 MB | 11.97 GB |
| | RoBERTa | 2,262 MB | 1.47 GB |
| | LayoutLM | 1,921 MB | 2.39 GB |
| Encoder-Decoder | BART | 1,578 MB | 2.50 GB |
| | MBART | 2,440 MB | 10.98 GB |
| | PEGASUS | 3,370 MB | 5.11 GB |
| Decoder | GPT2 | 12,101 MB | 0.30 GB |
| | GPT-NeoX | 57,800 MB | 2.25 GB |
| | OPT | 1,907 MB | 1.98 GB |
| | Ernie 3.0 | 2,234 MB | 0.324 GB |
| | BLOOM | 26,800 MB | 10.08 GB |
| | GLM | 26,800 MB | 36 GB |
| | Falcon | 100,710 MB | 18 GB |
| | Llama 3 | 16,700 MB | 126 GB |

When choosing an LLM from the table for text analytics, it is important to refer to the estimated disk space and GPU requirements. As LLM parameters increase, so do the disk space and estimated GPU requirements. Some models (e.g., Llama 3) are too large to be fine-tuned using consumer hardware. However, GPU requirements can be potentially reduced using quantization (bit reduction), batch size (number of samples per training iteration), and prompt tuning techniques [49].

## 4 LLMS FOR TEXT ANALYTICS INFORMATION SYSTEMS RESEARCH (TAISR) FRAMEWORK

To assist IS researchers in implementing LLM, we present a Text Analytics Information Systems Research (TAISR) framework (Figure 4) with five major components: (1) Research Objective, (2) Data Collection, (3) Text Analytics Task, (4) LLM Implementation, and (5) evaluation. Our proposed TAISR framework provides targeted recommendations for IS researchers to implement in various research genres. Although not all five components are necessary for every research project, the recommendations found within each component provide valuable and generalizable insights for incorporating LLMs into IS research.

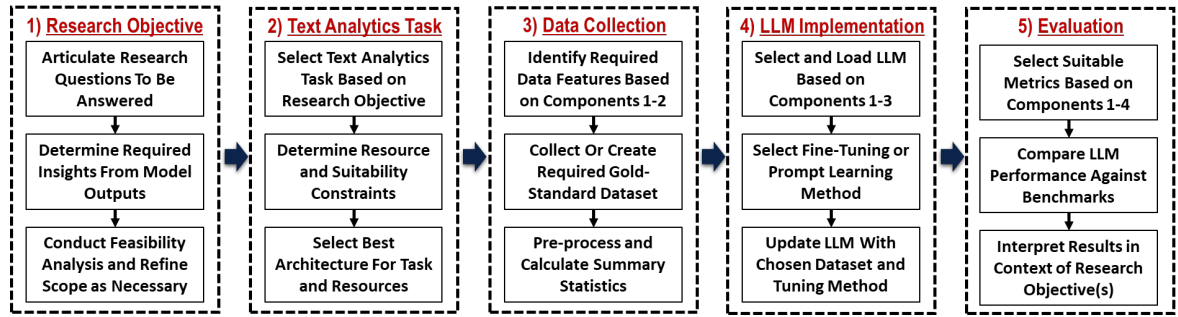

Figure 4: The Proposed TAISR Framework

### 4.1 Research Objective

The research objective provides initial guidelines that inform the rest of the LLM framework. AI systems have been used across each IS paradigm (design science, behavioral, econometrics) [63, 69, 70]. Design science research often focuses on the creation of AI artifacts for specific use cases [69]. Behavioral research often implements AI to generate variables [18] or obtain deeper insights into user-generated text [1]. Econometrics research often implements AI to understand phenomena through feature generation and improve causal inference [63]. An LLM solution within these three IS paradigms should still follow recommended IS practices [63, 69, 70]. Based on the literature and real-world insights, IS researchers should aim to develop questions that address novel and societally relevant problems [13, 29]. These questions should inform model requirements (what domain characteristics need to be captured by the LLM?) and necessary outputs (what LLM outputs will answer my questions?). Based on these requirements, researchers should conduct a feasibility analysis using small-scale data and classical machine learning models to determine whether an LLM is required. For example, advanced classical machine learning models (e.g., XGBoost) continue to achieve excellent performance in text classification tasks at a fraction of training time and cost [40]. The feasibility analysis can also include the operationalization of the kernel and design theories to further refine the LLM design requirements [29]. The results of the feasibility analysis should be used to iteratively update the research questions and scope as necessary.

### 4.2 Text Analytics Task

The text analytics task is selected based on the research objectives. Generally, LLMs are powerful tools for classification, summarization, and generation [57]. First, text classification tasks are frequently used to infer data labels. Design science research frequently aims to build classification models to achieve state-of-the-art inference or detection performance in various IS tasks [3, 4, 86]. Behavioral research can use text classification models to infer features (e.g., sentiment) used for downstream regression or structural equation modeling [18]. Econometric research can use classification models for explanatory variable feature prediction [63]. Second, text summarization tasks are frequently used to extract vital information from large and unstructured textual datasets (e.g., summarized reports from financial posts) [25]. Design science research can enhance text summarization models for targeted use cases. Behavioral research can benefit from text summarization principles for multi-lingual analysis (e.g., translating construct items from one language to another) or key insight extraction (summarizing groups of open-ended responses in qualitative research). Econometric research can summarize lengthy reports or articles, and then apply topic frequency models to create new and interesting explanatory variables. Finally, text generation tasks often aim to build or implement an artifact that can create open-ended text (e.g., chatbots and idea creation). Design science research can consider the methodological framework surrounding text generation models that lead to the best text for a target task. Behavioral research can consider analyzing user behavior with LLM-enabled chatbots [71] or using text generation methods for ontological mapping [22, 47]. Econometric research can incorporate econometric modeling results into text generation procedures [28].

The implementation of the three classes of tasks varies significantly in terms of data labels, quantity, and goals. IS researchers should first determine their computational resources. As shown in Table 4, LLMs with more parameters require more computational power. Large amounts of computing resources are not typically available in business schools. Therefore, IS researchers should consider smaller versions of seminal LLMs (e.g., TinyBERT [38]) or implementing efficient training strategies (e.g., low-rank adaptation [32]) depending on their computing resources.

Once the text analytics task and computing resources have been determined, researchers should select their LLM architecture. We recommend that IS research focuses on open-source LLMs. Closed-source LLMs are cost-prohibitive, not interpretable, rarely allow fine-tuning, and may retain proprietary data. In our opinion, the potential performance boost offered by these models is not worth these limitations. As stated previously, encoder LLMs (e.g., BERT) excel at classification, encoder-decoder LLMs (e.g., BART) excel at summarization, and decoder LLMs (e.g., GPT-2) excel at generation. However, the choice of LLM within these tasks is highly dependent on the language (e.g., English or multilingual), requirements (e.g., language reasoning or simple outputs), and GPU availability (e.g., high or low). To assist researchers in determining which LLM to select based on their task, we provide a flowchart in Figure 5.

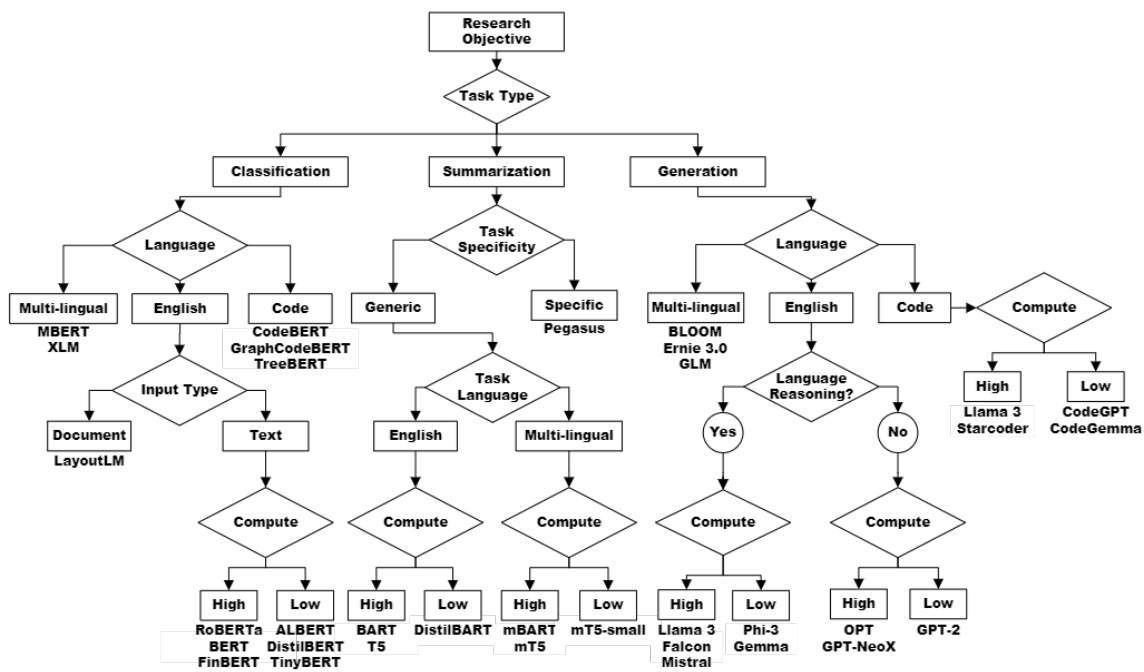

Figure 5: The TAISR Text Analytics Task Flowchart

Within classification, researchers should first consider what language they work with, as some models are multi-lingual (e.g., XLM), whereas others only process English text (e.g., BERT). Researchers should then consider whether they want to work with individual sentences or passages (e.g., sentiment analysis) or if they want to consider document structure as well (e.g., automated document-to-spreadsheet conversion). Finally, researchers must consider their computational power. LLMs for classification are generally less computationally intensive than their summarization and generation counterparts. Within summarization, researchers should first determine whether their research objective is generic (e.g., news articles or book summarization) or specific (industry-specific or jargon-laden domains). Models like Pegasus generalize well to specific domains with fine-tuning [90]. Like classification, summarization tasks have English-specific (BART) and multi-lingual-specific (mBART) models. Finally, summarization models also have high- and low-resource models. Within generation, researchers should first consider the type of language they would like to generate. Researchers should then consider whether language reasoning (the ability of the model to provide insight into why it generates a certain text) is important for their task. For example, chatbots can significantly benefit from language reasoning ability to reduce hallucinations [37]. Finally, researchers need to consider computing power. Text generation LLMs have the highest resource requirements. Llama 2 (and 3) require over 100GBs of virtual memory to fully fine-tune. While there are methods to reduce memory requirements (quantization, batch size, prompt tuning), researchers should consider if the additional performance between the high (Llama 3, Falcon) and low (Phi-3, Gemma) resource models are worth the cost.

Although Table 4 and Figure 5 provide examples of seminal open-source models in each LLM architecture, these recommendations will quickly become outdated. Researchers can discover new and popular LLMs for each task on Hugging Face, following top AI outlets (e.g., NeurIPS, ICLR, ICML, AAAI), and top AI research groups (e.g., Google, Meta) on LinkedIn.

### 4.3 Data Collection

Data collection requires obtaining a representative dataset that can address research question(s) and fulfill text analytics task requirements. The data requirements can also be determined from the feasibility analysis. These requirements can include textual structure (e.g., news articles have a title, author, tags, and main content) [25], user information (e.g., demographics of who is generating the textual content) [54], ground-truth (e.g., annotated labels, part-of-speech tags) [3], metadata (e.g., temporal or geospatial features) [81], and availability (e.g., potential dataset size and diversity).

Unless an IS researcher has access to proprietary organizational data or a representative dataset, they will have to collect their own textual data. There are currently two main methods for collecting large textual datasets within IS text analytics research: (1) open-source scraping and (2) crowdsourcing [12, 54]. First, open-source scraping involves identifying an online collection of textual data (e.g., blog posts), building a crawler to collect and parse it, and storing the parsed text. Open-source scraping has been implemented by IS research to collect blog/social media posts [12, 92], news articles [2], hacker communities [3, 24], and health information [81]. Scraping data from online communities is inexpensive, easy to scale, and is often how LLM creators gather pre-training datasets [20, 68]. However, open-source text is often observational, biased, and does not allow fine-grained control of the data collection. Second, crowdsourcing platforms (e.g., Mechanical Turk) are often used by the IS community to quickly mine specific opinions and behaviors [54]. Crowdsourcing platforms are more expensive to collect than open-source collections, but allow fine-grained insights [74]. For example, crowdsourcing platforms can collect demographic information, ask follow-up questions, and potentially detect disingenuous or biased answers [74, 83].

Once data is collected, it is important to conduct pre-processing and calculate summary statistics if necessary. Data pre-processing steps include manual labeling (e.g., classification and summarization tasks), cleaning (e.g., deduplication, formatting, and lemmatization), and debiasing (e.g., positive resampling) [9, 42]. These steps depend heavily on the research objectives and should be determined by data collection and the domain of interest. Manual feature engineering is not typical for LLMs because their self-attention mechanisms automatically determine salient textual features [67, 80]. After pre-processing, researchers calculated the summary statistics. This step will help determine the text analytics task. For labeled data, determining the distribution of labels assists in determining whether class imbalances must be accounted for [6]. For unlabeled text, summary statistics are less important. However, metrics like average document length, term frequency, and vocabulary complexity can assist in determining an appropriate LLM [88].

### 4.4 LLM Implementation

LLM implementation is where the core technical contributions of an IS research project can be made. Computational requirements will continue to increase and seminal LLMs will continue to change. Therefore, novel frameworks built upon LLM architectures (and not specific models) should have lasting impacts. While it is unlikely that IS will make new contributions on LLM pre-training (due to computational constraints), there is significant potential for novel technical contributions in building an LLM-based framework. We posit that technical novelties can be made in four key areas. First, novel textual input representations can be created prior to being input into an LLM. For example, Abbasi et al. [1] created a novel data representation using character, representation, demographic, and word embeddings from user surveys for enhanced psychometric measure analysis. While the authors input the embeddings into BiLSTM and CNN models, it would be trivial to update the BiLSTM with a BERT encoder to enhance the output of their psychometric analysis model. Second, the fine-tuning strategy can be updated based on task requirements and data availability. For example, Ampel et al. [4] formulated a knowledge distillation strategy to extract targeted information from the English-based RoBERTa model and the source code-based CodeBERT model for hacker exploit linking. Third, decoding methods for summarization and generation tasks are continuously updated and evaluated [45, 75]. The choice of decoding strategy has a significant impact on the quality of the generated text [31]. Therefore, a custom decoding strategy rooted in domain requirements can lead to significantly improved LLM output. Fourth, LLM outputs can be used as inputs in subsequent models. Research suggests that training a classification model with an LLM-generated synthetic dataset can significantly improve the model performance in data-sparse domains [56]. Behavioral and econometric research can consider synthetic dataset creation to enhance feature labeling in sparse domains (e.g., small-scale organizational or electronic health records). While we emphasize these areas of novelty, researchers should not feel constrained by them as LLMs continue to evolve rapidly.

To implement the selected LLM, researchers should use a virtual Python environment or Google CoLab. While the free version of Google CoLab has limited GPU resources, paid tiers provide limited access to high-end GPUs (e.g., NVIDIA A100). CoLab can be cost-effective for research projects that require one-time labeling (e.g., sentiment analysis tasks in [18]). For research that requires long-term LLM implementation (e.g., chatbots), IS researchers will need access to on-site GPU computing. Most open-source LLMs are available using Python. Using the Python environment of choice (e.g., Visual Studio, Anaconda, or Google CoLab), IS researchers can download their chosen LLM (usually through Hugging Face or OpenAI) and load their dataset (usually as a PyTorch dataset). The fine-tuning strategy (or lack thereof) should be based on LLM, text analytics tasks, and data collection.

For classification tasks (e.g., sentiment analysis), typically a feed-forward classification layer is placed at the end of the LLM and is trained the same as a standard classification model [57]. This final layer must be fine-tuned. However, LLM layers can be frozen (either all or just the lower layers) if the training dataset is not sufficiently large (the current rule of thumb is 1,000 data points per label), if the research team does have access to adequate computing resources, or if the new input data is already semantically similar to the LLM training dataset [51].

For summarization tasks (e.g., financial summaries), fine-tuning often involves a set of inputs and outputs (e.g., a news article and ground-truth summary) that are placed into the encoder and decoder, respectively [25]. The weights of both the encoder and decoder are iteratively updated through the training process. The choice of decoding strategy (deterministic, stochastic, contrastive) should also be carefully experimented with during the fine-tuning phase. If ground-truth summaries are not available, some decoder LLMs (e.g., Llama 3) can perform well in zero-shot settings (e.g., no model weight updating) with careful prompt engineering [79].

For generation tasks (e.g., chatbots), many open-source models work well in the zero-shot setting [49]. However, these models often produce incorrect information (known as hallucinations) if the research domain is not similar to the LLMs training dataset [37]. Generative LLMs can often be improved with autoregressive fine-tuning (updating LLM weights with a representative dataset), knowledge enhancement (providing a database the LLM can reference), and prompt learning (learning the best input prompt) [10, 33, 34].

### 4.5 Evaluation

Once the LLM has been adapted for the chosen text analytics task, it is vital to evaluate its performance. In design science, this evaluation is often performed against current state-of-the-art approaches [70]. In behavioral and econometric research, these evaluations can be as simple as the error rate to ensure that the implemented LLM achieves its target task (e.g., feature imputation or sentiment analysis). Our general recommendations for evaluating adapted LLMs follow the seminal literature on technical evaluations of machine and deep learning IS research [63, 70]. Evaluation metrics should be aligned with the previously established TAISR components (i.e., research objectives, text analytics task, dataset, and model architecture). IS researchers should consult recent and relevant literature to determine seminal metrics (e.g., Table 2) that are suitable for their context. For example, behavioral research on users interacting with LLM chatbots should consider human metrics to evaluate their LLM output quality [71, 75]. Comparing the adapted LLM against the current state-of-the-art benchmark models can provide a comprehensive understanding of the strengths and weaknesses of the adapted LLM. It is also recommended that IS researchers conduct a thorough sensitivity analysis of their LLM [70]. This process includes the use of various fine-tuning strategies, datasets, and LLM architectures to determine the technical components that contribute the most to overall performance. IS researchers may also consider creating distilled (i.e., smaller and highly targeted) versions of their adapted LLM for sensitivity analysis to determine the tradeoff between performance (in terms of metrics) and LLM size (in terms of parameters and GPU requirements) [4]. These distilled models may have significant impacts within mobile device and ecological IS research [22]. This framework does not provide recommendations for the situated implementation of a trained LLM. However, non-technical evaluations of an adapted LLM in its intended context should be conducted following IS best practices [60].

## 5   CASE STUDIES

As a proof-of-value for our TAISR framework, we present three brief case studies in business intelligence that follow our recommendations for using LLMs (classification, summarization, and generation) and are relevant to current organizational goals and IS research: (1) Sentiment Analysis for Organizations, (2) Automated Reporting of Competitor Actions, and (3) Social Media Content Generation for Organizations. The case studies apply the principles of our TAISR framework to show how IS researchers can design future textual research streams for advanced business intelligence using LLMs. While our case studies focused on general organizational contexts, we also foresee several domain-specific uses. For example, healthcare could benefit from electronic health record synthesis and cybersecurity professionals could benefit from proactive threat intelligence generation [24, 41].

### 5.1   Case Study 1: Sentiment Analysis for Organizations

Organizations are increasingly interested in measuring public opinion in real time. Recent IS research has identified sentiment analysis of open-source data (e.g., financial reports, social media, product reviews) as a powerful tool for assisting organizations in the automated analysis of public opinion [21, 92]. Therefore, our first case study examined how LLMs can be used in the context of organizational sentiment analysis.

#### 5.1.1   Research Objective

Organizations are becoming increasingly interested in analyzing public opinion on them. Public organizational opinions often come in the form of financial news, social media, and product reviews [12, 21, 92]. However, collecting a large amount of labeled public sentiment from three separate platforms can be cost-prohibitive. This cost necessitates a model that can automatically determine public sentiment about an organization given limited data. Therefore, our research question is:

- How can we develop a framework that can accurately determine public sentiment about an organization across financial news, social media, and online product reviews when provided minimal data?

In a small-scale feasibility analysis, we found that classical machine learning and deep learning models struggled to classify public sentiments across these three platforms. However, LLMs may be a suitable solution for classifying sentiments across different platforms with appropriate fine-tuning and guidance [51].

### 5.1.2  Text Analytics Task

Given our goal of predicting public sentiment and our access to ground-truth labels, we chose a classification approach. For the purpose of this case study, we used a low-resource machine with one 16GB NVIDIA 4080 GPU. Therefore, our approach required a method that can be fine-tuned with minimal data. Following the task analytics flow chart, we adapted the encoder LLM FinBERT because of its state-of-the-art classification results in organizational tasks [36]. This LLM is an adapted version of BERT, fine-tuned on financial analysis reports.

### 5.1.3  Data Collection

To analyze an organization's public sentiment, we require a ground-truth dataset from each platform (financial news, social media, and online product reviews). For financial news, we collected 4,840 labeled news articles about organizations manually annotated as positive, neutral, or negative sentiments by financial experts [55]. For social media, we collected 10,000 labeled posts (positive or negative) about organizations from X (formerly Twitter). For online product reviews, we collected 25,000 Amazon reviews and their associated ratings (i.e., 1, 2, 3, 4, or 5). Our overall ground-truth dataset is shown in Table 5.

Table 5: Sentiment Analysis Data Collection Summary

| Dataset | Size | Label Summary Statistics |
|---|---|---|
| **Financial News** | 4,840 | **Positive**: 59.4%, 1,363 records **Neutral**: 28.10%, 2,879 records **Negative**: 12.60%, 604 records |
| **Social Media** | 10,000 | 5,000 records for each rating (positive and negative) |
| **Product Reviews** | 25,000 | 5,000 records for each rating (1, 2, 3, 4, and 5) |

From Table 5, we noted that the financial news dataset is imbalanced, favoring neutral and positive sentiments over negative ones. However, we did not believe that the imbalance in the dataset necessitated special training measures. Our social media and product review datasets were evenly split across each label.

### 5.1.4  LLM Implementation

A fine-tuning method is required to adapt the FinBERT model to our sentiment analysis task. As previously stated, partial fine-tuning or prompt learning is preferable when the pre-trained model has already captured relevant features for the target task in its lower layers (i.e., FinBERT is already adapted for financial analysis). We adopted the Low-Rank Adaptation learning strategy because we had a small amount of labeled data [32]. We used an Anaconda virtual environment with Hugging Face installed to download and load FinBERT. We loaded our collected dataset and used Hugging Face's built-in Low-Rank Adaptation method. We used a low learning rate of 0.01 to ensure that our LLM weights did not significantly shift. We updated the LLM for three epochs, as this was when the performance stopped changing.

*5.1.5 Evaluation*

We compared our tuned-FinBERT (FinBERT adapted on our datasets) against the seminal machine (Decision Tree, KNN, SVM, Logistic Regression) and deep learning (CNN, LSTM, BiLSTM, BiLSTM with attention) sentiment classification models [5, 21, 92]. We also compared our tuned-FinBERT against TinyBERT (a small and distilled version of BERT) and BERT [38]. Each model was trained (or fine-tuned using Low-Rank Adaptation) using the respective datasets. The results were evaluated using the F1-score (commonly used to evaluate imbalanced sentiment analysis) [5]. The results for each model are shown in Table 6.

Table 6: Results for Case Study 1: Sentiment Analysis for Organizations

| Model Type | Model | Dataset | | |
|---|---|---|---|---|
| | | FN | SM | PR |
| Classical Machine Learning | Decision Tree | 55.69% | 60.03% | 28.80% |
| | KNN | 61.24% | 62.16% | 31.09% |
| | SVM | 73.79% | 77.06% | 48.46% |
| | Logistic Regression | 78.13% | 78.46% | 43.09% |
| Deep Learning | CNN | 83.95% | 77.94% | 46.01% |
| | LSTM | 75.72% | 78.29% | 49.87% |
| | BiLSTM | 73.86% | 78.21% | 50.93% |
| | BiLSTM w/ Attention | 83.42% | 79.08% | 51.67% |
| LLMs | TinyBERT | 70.72% | 74.00% | 44.12% |
| | BERT | 85.15% | **81.10%** | 52.92% |
| | Tuned-FinBERT | **91.34%** | 78.60% | **54.40%** |

Note: BiLSTM = Bidirectional LSTM, CNN = Convolutional Neural Network, FN = Financial News, KNN = K-Nearest Neighbors, LSTM = Long Short-Term Memory, PR = Product Reviews, SM = Social Media, SVM = Support Vector Machine.

The results of our experiment show that tuned-FinBERT led to state-of-the-art results for sentiment analysis of financial news (91.34%) and product reviews (54.40%), while achieving the third best results for social media (78.60%). FinBERT performing well for financial news is consistent with literature [36]. Tuned-FinBERT underperforming BERT for the social media dataset (78.60% vs. 81.10%) is also consistent, as social media text is often more jargon-laden than news articles for which FinBERT has been specially tuned. F1-scores for product reviews were generally low for all models. These results are most likely due to the five-class classification setting and biased data (i.e., a person may say something nice and give a two-star review, and vice versa). However, tuned-FinBERT was still able to discover latent patterns and outperformed all other benchmark models. It was disappointing that TinyBERT (distilled BERT) performed poorly in all settings. However, TinyBERT is a distilled model that we then fine-tuned. It would be interesting in future research to determine whether the results hold if the order of operations is reversed (i.e., fine-tune and then distill to create TinyFinBERT).

## 5.2 Case Study 2: Automated Reporting of Competitor Actions

For organizations to maintain a competitive advantage, it is vital that they generate business intelligence through careful analysis of open-source internet data [12]. Open-source data can contain consumer opinions and information about an organization and its competitors. Understanding competitor actions is vital to the health and success of an organization [16]. Therefore, automatically generating reports of external competitor actions is of great interest to several types of organizations. Our second case study examined how LLMs can be used in the context of automated report generation.

### 5.2.1  Research Objective

To assist organizations in quickly understanding competitor actions, we aim to automate the generation of reports by summarizing recent and relevant news articles about competitors in our manufacturing company. However, summarizing news articles in a manner that is relevant to organizations is currently unclear. Therefore, our research question is:

- How can we develop a framework that summarizes news articles about competitor actions in a business intelligence report?

In a small-scale feasibility analysis, we found that seminal machine and deep learning models are currently not well prepared to generate high-quality summaries for specific contexts (i.e., they only perform well in general and high-data settings) [89]. LLMs have recently achieved state-of-the-art performance in many text summarization tasks, justifying them for our problem domain context [25].

### 5.2.2  Text Analytics Task

Given our goal of generating business intelligence from news articles, our text analytics task is a summarization approach. Summarization can be extractive (extracting keywords or sentences) or abstractive (generating summarized content) [25]. Extractive summarization is useful in the absence of ground-truth summaries. However, extractive summarization is often incoherent [48]. Abstractive summarization often creates the most coherent summaries if a representative ground-truth dataset is available for fine-tuning [46, 90]. Like Case Study 1, we used a machine with a 16GB NVIDIA 4080 GPU. Given our reporting generation goal, we followed the task analytics task flowchart and determined that Pegasus was appropriate for our summarization task, as it was pre-trained specifically for abstractive summarization [90].

### 5.2.3  Data Collection

Our business intelligence summarization task required two data features. First, we require news articles on competitors' actions. Second, we require ground-truth summaries of these news articles to help fine-tune an LLM. For this case study, we assumed the position of a researcher for a Fortune 500 consumer electronics company. Using open-source scraping, we collected 150 news articles about competing companies. We then manually crafted brief ground-truth summaries of each news article regarding the actions that we thought would be interesting to our company. In our dataset, each row corresponds to a paragraph in a news article and it's manually crafted summary. An example row of the dataset is shown in Table 7.

Table 7: Example News Article and it's Ground-truth Summary

| News Article Paragraph | Ground-truth Summary |
|---|---|
| South Korean electronics giant Samsung has made significant progress since its foray into the automotive device market in 2015. Recent launches of various automotive ICs showcased Samsung's effort to diversify business lines from the volatile consumer memory market and to create valid growth opportunities for its semiconductor foundry business. It aims to become the world's leading automotive memory company by 2025, surpassing Micron, the current champion with a 45% global market share. | Samsung has recently launched initiatives to become the world's leading automotive memory company, shifting focus away from consumer products. |

*5.2.4    LLM Implementation*

Given our ground-truth dataset and Pegasus' low GPU requirements, we conducted a full fine-tuning procedure. Like Case Study 1, we used an Anaconda virtual environment with Hugging Face installed to download and load Pegasus. We loaded our ground-truth news articles and associated summaries and fine-tuned Pegasus using Hugging Face's built-in modules. We used a low learning rate of 0.01 and updated the model for five epochs, as this is when the LLM performance stopped changing.

*5.2.5    Evaluation*

We compared our tuned-Pegasus against seminal deep learning (CNN, RNN, LSTM) and LLMs (BERT, BART) for summarization [43, 46, 52, 90]. Classical machine learning models are not often used for summarization due to the encoder-decoder requirement. The summarization results were evaluated using ROUGE-1 (measuring overlapping unigrams), ROUGE-2 (measuring overlapping bigrams), and ROUGE-L (measuring longest common subsequence) scores, which are common metrics in reference-based summarization tasks [90]. Each metric was measured between 0 and 1, with scores closer to 1 being better. The results for each model are shown in Table 8.

Table 8: Results for Case Study 2: Automated Reporting of Competitor Actions

| Model Type | Model | ROUGE-1 | ROUGE-2 | ROUGE-L |
|---|---|---|---|---|
| **Deep Learning** | CNN | 0.398 | 0.152 | 0.319 |
| | RNN | 0.372 | 0.174 | 0.375 |
| | LSTM | 0.401 | 0.171 | 0.394 |
| **LLM** | BERT | 0.384 | 0.164 | 0.366 |
| | BART | 0.427 | 0.208 | 0.409 |
| | Pegasus | 0.430 | 0.217 | 0.411 |
| | Tuned- Pegasus | **0.443** | **0.229** | **0.423** |

Note: BART = Bidirectional and Auto-Regressive Transformers, BERT = Bidirectional Encoder Representations from Transformers, CNN = Convolutional Neural Network, LSTM = Long Short-Term Memory, PEGASUS = Pre-training with Extracted Gap-sentences for Abstractive Summarization, RNN = Recurrent Neural Network.

The results of our experiment show that, among the deep learning models, the recurrent-based LSTM achieved the best ROUGE-1 and ROUGE-L scores. This result makes sense because the sequential and long-term memory features of LSTM are well equipped for summarization tasks. Among the LLMs, BERT performed the worst in all three metrics (and was outperformed by the LSTM model). These results are consistent with our literature review showing that encoder LLMs (i.e., BERT) are not appropriate for text generation or summarization tasks [57]. BART and Pegasus outperformed LSTM in all three metrics. These results suggest that their pre-training leads to significant improvements in text summarization. BART is not specifically conditioned for abstractive summarization, potentially explaining why Pegasus outperformed BART. Finally, fine-tuning Pegasus on our small ground-truth dataset led to state-of-the-art results in ROUGE-1 (0.443), ROUGE-2 (0.229), and ROUGE-L (0.423). It is important to note that ROUGE only assesses content selection based on potentially subjective ground-truth summaries. Therefore, IS researchers should also consider post-hoc human evaluations to gauge the fluency, coherence, and value of the generated business intelligence reports.

### 5.3    Case Study 3: Social Media Content Generation for Organizations

Social media interactions have become vital for the dissemination of organizational news, collecting opinions, increasing brand awareness, and building sustained relationships with customers [2, 17, 92]. However, the scale of social media makes it difficult to monitor and create content continuously. Therefore, our final case study examined how LLMs can automatically generate textual content for social media platforms.

### 5.3.1  Research Objective

The goal of Case Study 3 was to automate the generation of social media content based on a prompt. These social media posts must match the writing style of the organization to maintain brand coherence. Therefore, our research question is:

- How can we automatically generate social media content that matches the writing style of an organization based on existing social media interactions?

Like text summarization, machine and deep learning models are not well equipped to generate free-form text due to their low number of parameters and forgetting over long sequences [57]. Autoregressive LLMs (e.g., ChatGPT) currently provide state-of-the-art results in text generation, and should be investigated to answer our research question [61].

### 5.3.2  Text Analytics Task

Given our goal of generating open-ended social media posts, our text analytics task was a generation approach. Text generation is highly variable and depends on fine-tuning, prompting, and decoding [51, 57]. Similar to prior case studies, we used a low-resource machine with a 16GB NVIDIA 4080 GPU. We followed the text analytics task flowchart and determined that the task did not require language reasoning. Therefore, we chose the GPT-NeoX model as our LLM to implement [8].

### 5.3.3  Data Collection

To generate novel social media content, a representative dataset of real social media posts is required. For this case study, we chose the fast-food company Wendy's. Wendy's social media posts have recently become popular, frequently injecting humor and weirdness into their textual content as a strategy for increasing brand awareness [17]. Using open-source scraping, we collected 500 of Wendy's most recent replies to consumers on X (formerly Twitter). We also collected the posts that Wendy's was replying to. This allows us to use those posts as input prompts (i.e., "Generate a response to this post: [post content here]"). This dataset provides a representative sample of the type of textual content Wendy's is likely to post in response to consumer mentions of them.

### 5.3.4  LLM Implementation

Based on other text generation tasks, GPT-NeoX is the best-performing autoregressive LLM that we can reasonably fine-tune on our hardware (i.e., Llama 3 requires significantly more computational resources to fine-tune or various tricks to load into memory). Given our ground-truth dataset, we conducted an autoregressive fine-tuning procedure. We used an Anaconda virtual environment with Hugging Face installed to download and load GPT-NeoX and our Wendy's X dataset. We fine-tuned the parameters of GPT-NeoX using an autoregressive strategy (GPT-NeoX predicts the likelihood of the next token given the preceding Wendy's post context). We used a low learning rate of 0.01 and updated the model for three epochs (as this is when LLM performance stopped changing).

We compared GPT-NeoX against seminal text generation models (GPT-2, BLOOM, OPT, GPT-NeoX) [8, 44, 68]. Each benchmark model was fine-tuned in the same manner as that for GPT-NeoX. Among the decoding methods, we predicted that contrastive search would lead to the most natural-looking social media posts because it attempts to balance both diversity and coherence of text [75]. However, we also measured the performance of the top deterministic (greedy, beam) and stochastic (top-$k$, nucleus) decoding methods to further explore the effect of decoding strategy on text generation. We fine-tuned each LLM and prompted it to generate a response to the consumer posts in our Wendy's dataset. We calculated the perplexity, diversity, and coherence of the generated social media posts, as these are often used metrics in open-ended generation [31, 45, 75]. We also calculated these metrics on the real Wendy's responses. Our goal was not to find the lowest perplexity or highest diversity/coherence. Rather, the goal was to minimize the delta ($\Delta$) between the metrics of the real Wendy's replies and the LLM-generated replies to ensure that our responses are statistically representative of Wendy's style. The results for each model, decoding strategy, and method are listed in Table 9.

Table 9: Results for Case Study 3: Social Media Content Generation for Organizations

| Model | Decode Strategy | Decode Method | Perplexity | Diversity | Coherence |
|---|---|---|---|---|---|
| **Wendy's Replies** | N/A | N/A | 81.63 | 0.8614 | 0.6055 |
| **GPT-2** | Deterministic | Greedy | 159.1 (77.47) | 0.3962 (0.4652) | 0.5507 (0.0548) |
| | | Beam | 152.9 (71.27) | 0.4026 (0.4588) | 0.5567 (0.048) |
| | Stochastic | Top-k | 157.0 (75.37) | 0.6780 (0.1834) | 0.5150 (0.0905) |
| | | Nucleus | 116.5 (34.87) | 0.6566 (0.2048) | 0.5334 (0.072) |
| | Contrastive | Search | 110.6 (28.97) | 0.7479 (0.1135) | 0.6010 (0.004) |
| **BLOOM** | Deterministic | Greedy | 262.4 (180.77) | 0.4930 (0.3684) | **0.6045 (0.001)** |
| | | Beam | 236.0 (154.37) | 0.4766 (0.3848) | 0.6192 (0.013) |
| | Stochastic | Top-k | 234.4 (152.77) | 0.7252 (0.1362) | 0.6041 (0.0014) |
| | | Nucleus | 251.3 (169.67) | 0.7088 (0.1526) | 0.5926 (0.0129) |
| | Contrastive | Search | 227.6 (145.97) | 0.6883 (0.1731) | 0.5912 (0.014) |
| **OPT** | Deterministic | Greedy | 298.9 (217.27) | 0.4739 (0.3875) | 0.5482 (0.0573) |
| | | Beam | 228.6 (146.97) | 0.4498 (0.4116) | 0.5612 (0.0443) |
| | Stochastic | Top-k | 212.3 (130.67) | 0.7338 (0.1276) | 0.5932 (0.012) |
| | | Nucleus | 179.4 (97.77) | 0.7806 (0.170) | 0.6014 (0.004) |
| | Contrastive | Search | 117.6 (35.97) | 0.7551 (0.1063) | 0.5358 (0.0697) |
| **GPT-NeoX** | Deterministic | Greedy | 266.0 (184.37) | 0.449 (0.4124) | 0.5945 (0.011) |
| | | Beam | 216.0 (134.37) | 0.4822 (0.3792) | 0.5637 (0.041) |
| | Stochastic | Top-k | 177.0 (95.37) | 0.7253 (0.1361) | 0.6112 (0.005) |
| | | Nucleus | 104.4 (22.77) | 0.7378 (0.1236) | 0.5324 (0.073) |
| | Contrastive | Search | **90.3 (8.67)** | **0.7812 (0.0802)** | 0.5585 (0.047) |

Note: BLOOM = BigScience Large Open-science Open-access Multilingual Language Model, GPT = Generative Pre-trained Transformer, OPT = Open-source Pre-trained Transformer.

The results of our experiment show that GPT-NeoX outperforms other benchmark models in terms of perplexity and diversity. GPT-2 is an older autoregressive LLM with fewer parameters than modern LLMs, potentially limiting its ability to generate coherent and diverse text. BLOOM is a newer LLM trained on a massive corpus of multilingual data. Although BLOOM achieved the lowest difference in coherence when using greedy search ($\Delta$ 0.001), it struggled to develop diverse social media posts across all decoding methods. OPT, trained on a large corpus of internet content (e.g., Reddit posts, blogs), performs better than GPT-2 and BLOOM in diversity but struggles to generate coherent posts. Finally, GPT-NeoX (trained with similar internet content as OPT) achieved the lowest perplexity ($\Delta$ 8.67) and coherence ($\Delta$ 0.0802) difference when using contrastive search. Our evaluation also suggests that contrastive search outperforms deterministic and stochastic strategies in generating human-like X posts. However, human evaluations can further confirm the results of our analysis.

## 6 LLM CHALLENGES AND LIMITATIONS

Despite the potential usefulness of LLMs in conducting TAISR, they have several notable limitations that should be carefully considered before implementation in an organizational context [22]. We discuss some noteworthy limitations and potential mitigation strategies below.

### 6.1 Hallucinations

Hallucinations occur when LLMs provide falsified facts, code, references, or other information to a user [37]. Hallucination generally stems from LLMs that are trained on internet data that has not been cleaned or verified (e.g., social media posts) [23]. There are two types of LLM hallucinations: (1) intrinsic and (2) extrinsic [37]. First, intrinsic hallucinations are non-factual statements within LLM-generated texts. Intrinsic hallucinations can be mitigated by referencing external knowledge bases or by negative sampling. Second, extrinsic hallucinations are LLM-generated text that are not contextually relevant to the input prompt. RLHF has been shown to significantly improve open-ended LLM generation and reduce extrinsic hallucinations [62, 79].

### 6.2 Sensitive Information Disclosure

Sensitive information (e.g., personally identifiable information) is often learned by LLMs because it is trained on massive quantities of internet text [30]. This problem can be exacerbated by closed-access models (e.g., ChatGPT), which save user chat logs for future training instances. Sensitive information can be extracted from LLMs through targeted sampling of high-perplexity text sequences (which are often memorized training data) to reconstruct original training sentences [11]. Open-source LLMs also suffer from sensitive information disclosure owing to the severe vulnerabilities introduced by poor open-source coding practices [39]. To mitigate sensitive information disclosure, LLMs should be fine-tuned with differential privacy techniques or pre-trained with data that is scrubbed of sensitive information [53].

### 6.3 Detecting LLM-generated Text

LLM-generated text is a rising concern because it can be used for academic dishonesty, fraud, and misinformation [73]. Organizations have focused on developing detectors to automatically detect LLM-generated texts. Generally, there are three types of LLM detectors: (1) perturbation, (2) adversarial, and (3) classifier. First, the suspected LLM-generated text can be randomly perturbed and an LLM is asked to regenerate the text. Log probabilities are then calculated to determine the similarity between texts [87]. Second, a binary classifier can be jointly trained with an adversarial paraphraser to detect LLM-generated texts [35]. Third, a machine or deep learning model for binary classification (real or LLM-generated) can be trained [14]. Although these models can work within specific settings, they are highly sensitive to human paraphrasing, non-English text, and stochastic decoding methods [58]. Therefore, researchers should be wary of implementing LLM-generated text detection systems, and several potential research streams remain for improving detection methods.

### 6.4 Adversarial Prompts

Many LLMs have implemented security measures to prevent model misuse (e.g., sensitive information disclosure attacks) [61]. However, LLM security measures can be bypassed using adversarial prompts [19]. Adversarial prompts attempt to induce model misuse through the careful manipulation of an input prompt using competing objectives or mismatched generalization [93]. Competing objectives surround a malicious goal (e.g., write a phishing email) with benign text (e.g., generate a fairy tale story). This may cause misclassification by LLM security measures, leading to the generation of both benign and malicious objectives. In mismatched generalization, an LLM is presented with input data unknown to the LLM security measure (e.g., random characters, a different language). It is recommended to include input sanitization procedures and place an adversarial prompt classifier between the input and downstream LLM to help protect against adversarial prompting [19].

## 7 CONCLUSION

LLMs are rapidly becoming ubiquitous in text analytics research. However, there is not currently a framework to guide IS research in implementing LLMs for organizational contexts. Therefore, we proposed a TAISR framework comprising five components. We also provided three case studies to demonstrate the proof-of-value of our proposed framework. IS researchers can expand their research toolbox to include state-of-the-art LLMs by following our TAISR framework. While the TAISR framework is not comprehensive for all scenarios or LLM implementation methods, it provides a starting block for building interesting new research streams. We believe that these new LLM-enabled research streams will lead to interesting and novel insights into new and existing organizational problems.